%% file: main.tex
\documentclass{article}

\usepackage{arxiv}

\usepackage[utf8]{inputenc} 
\usepackage[T1]{fontenc}    
\usepackage{hyperref}       
\usepackage{url}            
\usepackage{booktabs}       
\usepackage{amsfonts}       
\usepackage{nicefrac}       
\usepackage{microtype}      
\usepackage{lipsum}
\usepackage{multirow}
\usepackage{algorithm}
\usepackage{algpseudocode}
\usepackage{amsmath}
\usepackage{array, makecell}
\usepackage{graphicx}
\usepackage{soul,color}
\usepackage{xcolor}
\usepackage{tabu}

\usepackage{subcaption} 

\usepackage{wrapfig}
\usepackage[draft,inline,nomargin,noindex]{fixme}

\title{Effective and Efficient Computation with Multiple-timescale Spiking Recurrent Neural Networks}

\author{
  Bojian Yin\\
  CWI \\
  \texttt{Bojian.Yin@cwi.nl} \\
   \And
Federico Corradi \\
  IMEC\\
  \texttt{Federico.Corradi@imec.nl} \\
  \And
  Sander M. Boht\'e\\
  CWI \\
  \texttt{S.M.Bohte@cwi.nl} \\
}

\begin{document}
\maketitle

\begin{abstract}
The emergence of brain-inspired neuromorphic computing as a paradigm for edge AI is motivating the search for high-performance and efficient spiking neural networks to run on this hardware. However, compared to classical neural networks in deep learning, current spiking neural networks lack competitive performance in compelling areas. Here, for sequential and streaming tasks, we demonstrate how a novel type of adaptive spiking recurrent neural network (SRNN) is able to achieve state-of-the-art performance compared to other spiking neural networks and almost reach or exceed the performance of classical recurrent neural networks (RNNs) while exhibiting sparse activity. From this, we calculate a $>$100x energy improvement for our SRNNs over classical RNNs on the harder tasks. To achieve this, we model standard and adaptive multiple-timescale spiking neurons as self-recurrent neural units, and leverage surrogate gradients and  auto-differentiation in the PyTorch Deep Learning framework to efficiently implement backpropagation-through-time, including learning of the important spiking neuron parameters to adapt our spiking neurons to the tasks.
\end{abstract}


\input{2_Introduction.tex}

\input{3_RelatedWork.tex}

\input{4_SRNN_training.tex}
\input{5_results.tex}

\input{Energy.tex}

\input{6_Conclusions.tex}

\bibliographystyle{unsrt}  
\bibliography{egbib.bib}  

\end{document}

%% file: 2_Introduction.tex
\vspace{-0.2cm}
\section{Introduction}
Modern deep neural networks have become highly capable in select applications of artificial intelligence. However, despite their effectiveness, their energy consumption is a limiting factor for application in many always-on application scenarios, like wearable intelligence devices, surveillance camera's and smartwatch applications \cite{Roy2019-rv}.
Standard efficiency approaches reduce the bit-precision of weights \cite{courbariaux2015binaryconnect,gong2019differentiable} and activations \cite{rastegari2016xnor,darabi2018bnn+}, or scale and prune models \cite{tan2019efficientnet,yang2017designing}. These methods however still adhere to the standard model of computation in artificial neural networks (ANNs), where activations are exchanged between neurons in a synchronous frame-based manner at every iterative processing step.

Taking inspiration from the extremely efficient brain, spiking neural networks (SNNs) \cite{maass1997networks} combine binary valued activations (spikes) with asynchronous and sparse communication. Such SNNs are arguably also more hardware friendly \cite{davies2018loihi} and energy-efficient \cite{panda2019towards}. However, compared to ANNs, the development of SNNs is in its early phase. Training deep SNNs has remained a challenge as the spiking neurons' activation function is typically non-differentiable and SNNs are thus not amenable to standard methods of error-backpropagation \cite{bohte2011error,neftci2019surrogate}. In particular, as spiking neurons can be modeled as a class of self-recurrent neurons, learning algorithms have to account for the past, and the resultant statefulness of SNNs makes it hard to deal with simulating and training very large networks. Finally, many deep learning benchmarks are geared towards the synchronized and iterative processing paradigm of ANNs, exemplified by image classification tasks.


Recent work \cite{neftci2019surrogate, bohte2000spikeprop} has demonstrated how the problem of a discontinuous gradient in spiking neural networks can be overcome effectively in a generic fashion through the use of {\em surrogate gradients}. This opens up new opportunities to leverage mature deep learning techniques in larger and more complex SNNs. 

Here, we develop compact recurrent networks of spiking neurons (SRNNs) which we train using such surrogate gradients to directly apply back-propagation-through-time (BPTT) with auto-differentiation in a well-developed modern deep learning framework (PyTorch). Using this framework, we can also easily train the parameters of the spiking neurons themselves, also for complex spiking neuron models with multiple dynamical timescales. As we show, this approach makes it feasible to adapt such spiking neural networks to the particular temporal dynamics of the task. 

We focus on sequential and streaming classification benchmarks with limited input dimensionality, including the well known sequential and permuted-sequential MNIST datasets (S-MNIST, PS-MNIST), the QTDB waveform classification of ECG, and the audio Spiking Heidelberg Digits dataset (SHD), tasks exemplary for various always-on edge computing devices that require low-power consumption. We demonstrate how our compact SRNNs can solve these complex tasks, exceeding SoTa in SNNs, and approaching or even exceeding SoTa compared to classical ANNs. On these tasks, the SRNNs demonstrate low to very low sparsity, and we show that this results in an $>$100x improvement in theoretical energy use over high-performing ANNs.

%% file: 3_RelatedWork.tex
\vspace{-0.1cm}
\section{Related work}
In standard deep learning, convolutional neural networks (CNNs) are widely used on visual tasks such as image classification and object recognition, while recurrent neural network (RNNs) are more generally applied to tasks that involve temporal patterns fed into the network as sequential input. In RNNs,  recurrency in the network induces memory in the form of internal hidden states $h_t$ which are updated while time-stepped input $x_t$ feeds in. 
For learning, because of the induced memory, RNNs are typically unrolled in time, for example using Backpropagation-Through-Time (BPTT) \cite{werbos1990backpropagation,mozer1995focused} to account for the relationship between past inputs and current state. BPTT however is both computationally expensive and tends to suffer from stability problems when computing the gradients.

Several alternative RNN variants have been developed to ease and improve learning in standard RNNs. The LSTM (Long Short-Term Memory) unit \cite{hochreiter1997long} was designed specifically as an RNN for sequential machine learning tasks like speech recognition, language understanding and sequence to sequence translation \cite{graves2013generating}. More recent innovations, like the IndRNN \cite{li2019deep}, borrow from the success of residual connections in CNNs to facilitate the gradient flow in the network and achieve state-of-the-art RNN performance. Alternatively, causal convolutional neural networks \cite{oord2016wavenet} have also been applied successfully to sequential tasks \cite{benidis2020neural} but have substantial network-size and data-history memory requirements.

Spiking neural networks \cite{maass2001pulsed} comprise of a class of event-based neural networks inspired by more detailed models of biological neurons. Biological neurons differ from the standard neurons in ANNs in the sense that they have internal state and communicate via isomorphic electrical pulses - spikes. The low average firing rate in the brain [1-5Hz] \cite{attwell2001energy} suggests that much effective and efficient computing can be done with stateful event-driven neurons that only sparingly exchange binary values \cite{panda2019towards}.  

For SNNs, learning rules for both feedforward and recurrent spiking neural networks have been developed \cite{bohte2011error,bohte2000spikeprop,bellec2018long,zenke2018superspike,Shrestha2018,huh2018gradient,ponulak2008analysis}, applying different types of spike-coding paradigms and learning methodologies. Recent work has achieved high performance in tasks like image classification \cite{zambrano2016fast,tavanaei2019deep}; still, it is unclear whether such SNNs are more efficient compared to conventional CNNs.  

One direction where potentially a clear advantage for SNNs can be obtained is tasks that fit their inherently temporal mode of computation and that can be computed with relatively compact networks fitting low-power neuromorphic hardware. Recent work has shown substantial progress in LIDAR \cite{Wang2020-uq} and speech recognition \cite{bellec2018long,cramer2019heidelberg}.
Still, in these tasks, a significant performance gap exists between SNNs and current deep learning solutions. 

\begin{figure}
  \begin{subfigure}[t]{0.475\textwidth}
    \includegraphics[width=\textwidth]{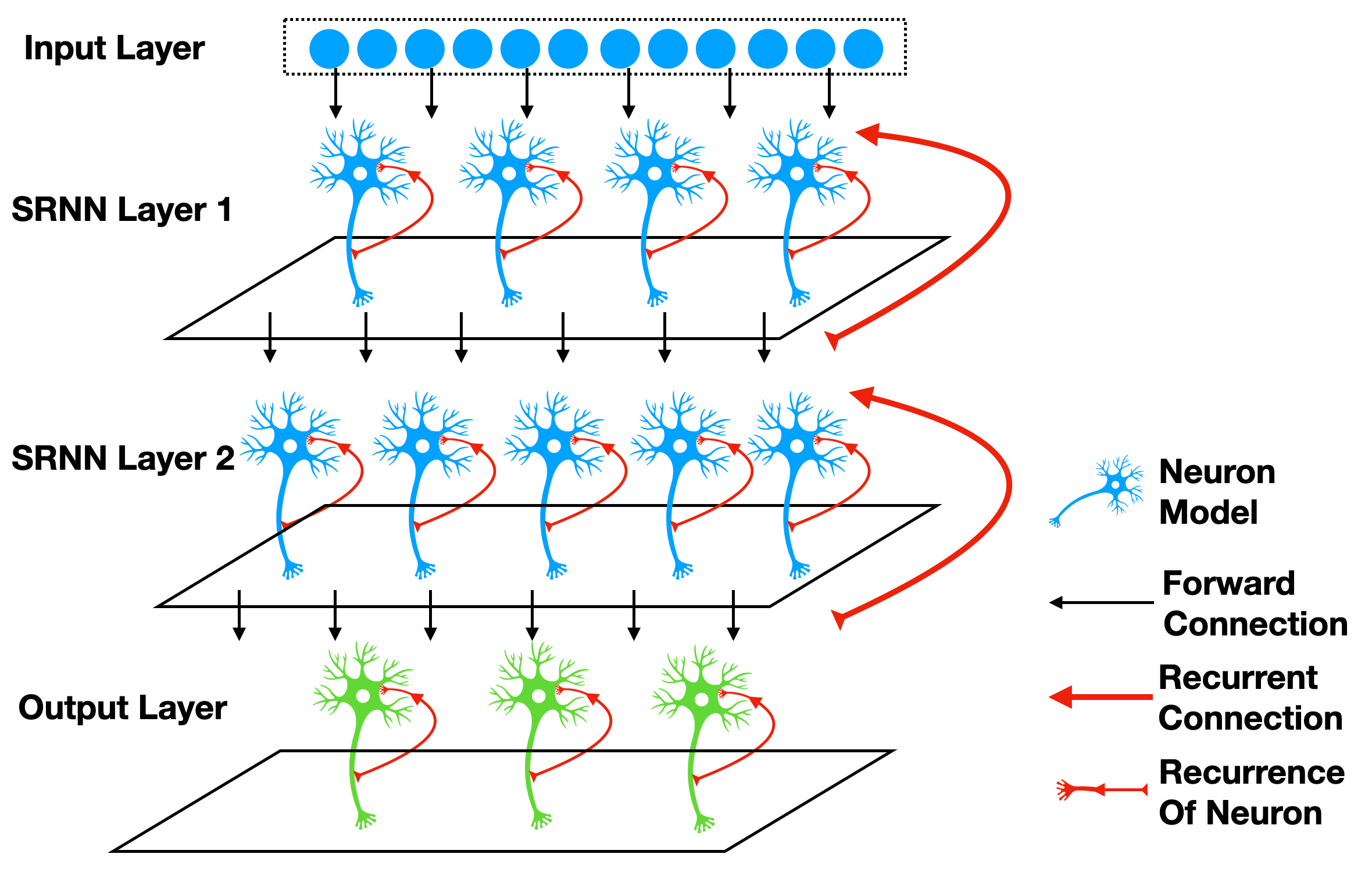}
    \caption{An SRNN with two recurrent layers. Neurons within a layer are fully recurrently connected, between layers neurons are fully connected with forward connections. }
    \label{fig:srnn_example}
  \end{subfigure}
  \quad
  \begin{subfigure}[t]{0.475\textwidth}
    \includegraphics[width=\textwidth]{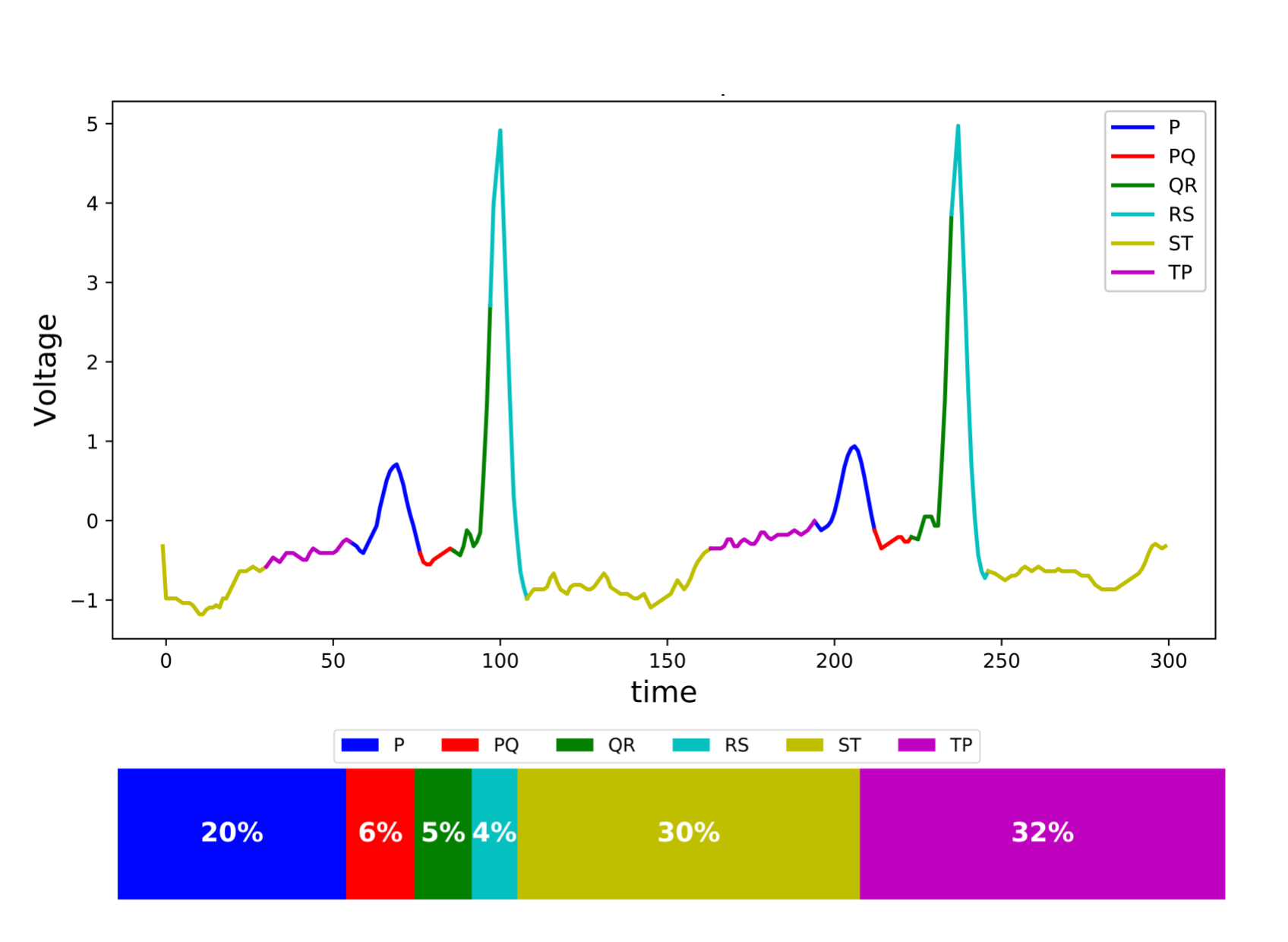}
    \caption{Top: example of the labeled ECG signal. The six basic parts of the signal, P,T and complex Q,R,S wave are color labeled. Bottom: relative waveform label distribution. }
    \label{fig:dataset_ecg}
  \end{subfigure}

\end{figure}

%% file: 4_SRNN_training.tex

\section{Spiking Recurrent Neural Networks}

Here, we focus on SNNs that comprise of one or more recurrent layers, Spiking Recurrent Neural Networks (SRNNs), illustrated in Fig. \ref{fig:srnn_example}. Within these networks, we  use one of two types of spiking neurons: Leaky-Integrate-and-Fire (LIF) neurons and Adaptive spiking neurons. Spiking neurons are derived from models that capture the behavior of biological neurons \cite{Gerstner2002-wd}. Complex models like the Hodgkin-Huxley model describe the detailed dynamics of biophysical quantities but are costly to compute; phenomenological models like the Leaky-Integrate-and-Fire (LIF) neuron model or the Spike Response Model trade-off levels of biological realism for interpretability and reduced computational cost.

\begin{figure}[t!p]
    \centering
    \vspace{-0.05cm}
    \includegraphics[width=\textwidth]{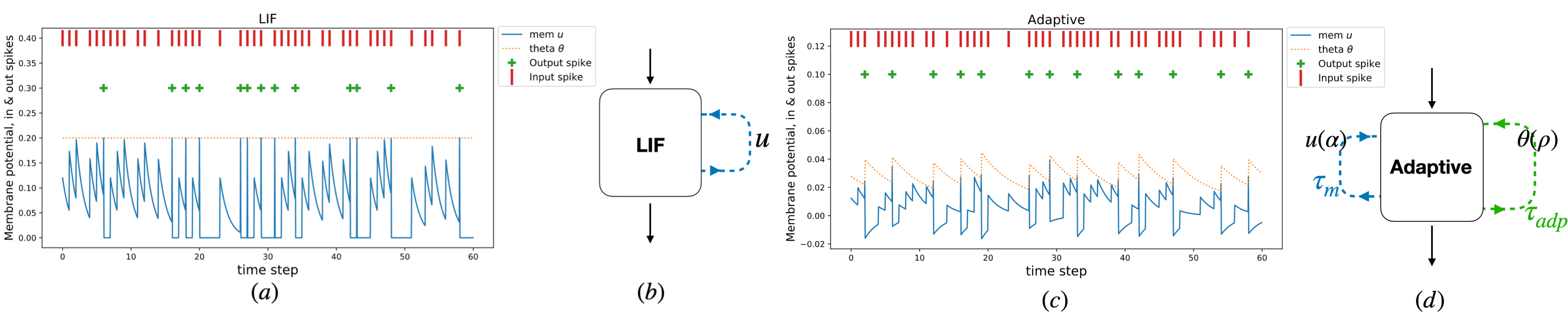}
    \vspace{-0.25cm}
    \caption{LIF and Adaptive spiking neuron behaviour. (a,c) Impinging input spikes (red) increase (or decrease) the membrane potential (blue) which then decays back to the resting potential (0). When the potential reaches the fixed threshold $\theta_0$ (yellow dotted line), an output spike is emitted (green cross) and the potential is reset to the resting potential. In the Adaptive spiking neuron (c), the threshold is increased for every emitted spike, and then decays back to the resting threshold $\theta_0$. (b,d) The decay of the membrane potential and adaptation can be modeled as self-recurrency. 
}
\vspace{-0.25cm}
\label{fig:neurons_behavior}
\end{figure}

{\bf The LIF spiking neuron} integrates input current $I_t$ in a leaky fashion and fires an action potential when its membrane potential $u_t$ crosses a fixed threshold $\theta$ from below, at which time a spike $s_t$ is emitted, a process modeled as a nonlinear function $\hat{f}_s(u_t,\theta)$, after which the membrane potential is reset to $u_r$:
\begin{align}
\label{eq1:LIF}
&\tau_m \frac{du}{dt} = -(u_t - u_r) + R_m I_t \\
& s_t = \hat{f}_s(u_t,\theta) \\
& u_t = u_t(1-s_t) + u_r s_t,
\end{align}
where $I_t = \sum_{t_i} \delta(t_i)$ is the input signal expressed as a spike-train $\{t_i\}$, $u_t$ is the membrane potential decaying exponentially with time-constant $\tau_m$, $R_m$ is the membrane resistance, and the emission of spike $S_{pike}$ is expressed as a nonlinear function of threshold and potential: $S_{pike} = \hat{f}_s(u_t,\theta)$. The LIF neuron is cheap to compute \cite{izhikevich2003simple}, but lacks much of the more complex behavior of real neurons, including responses that exhibit longer history dependencies. 

Bellec et al \cite{bellec2018long} demonstrate that using more complex, adapting spiking neuron models improved performance in their SNNs. In the adaptive spiking neuron, the threshold is increased after each emitted spike and then decays exponentially with time-constant $\tau_{adp}$. Simulating the continuous neuron model in discrete time using the forward-Euler first-order exponential integrator method for $dt=1ms$, we compute:
\begin{align}
\label{eq2:adapt}
&\alpha = \exp(-dt/\tau_m) \\
&\rho = \exp(-dt/\tau_{adp}) \\
&\eta_t = \rho \eta_{t-1}+ (1-\rho)S_{t-1} \\
&\theta = b_0+\beta\eta_t \\
&u_t = \alpha u_{t-1} + (1-\alpha)R_m I_t - S_{t-1}\theta,
\label{eq2:adaptsn}
\end{align}
where $\theta$ is a dynamical threshold comprised of a fixed minimal threshold $b_0$ and an adaptive contribution $\beta \eta_t$; $\rho$ expresses the decay single-timestep decay of the threshold with time-constant $\tau_{adp}$. The parameter $\beta$ is a constant that controls the size of adaptation of the threshold (eq. \eqref{eq2:adaptsn}); we set $\beta$ to 1.8 for adaptive neurons as default. 
Similarly, $\alpha$ expresses the single time-step decay of the membrane potential with time-constant $\tau_m$.

Fig.~\ref{fig:neurons_behavior} illustrates the behaviour of the two spiking neuron models in terms of evolution of the membrane potential, threshold and spiking behavior. Inspecting these neuron models, we see that the evolution of the membrane potential is determined by the self-decay term $\alpha$ and similarly, for the adapting neuron, the threshold by the self-decay term $\rho$: effectively, the behavior of spiking neurons can be modeled as being self-recurrent with weights $\alpha$ and $\rho$ (Fig.~\ref{fig:neurons_behavior}). In our network implementation, we set these self-recurrent parameters, $\tau_m$, $\tau_{adp}$, as trainable, as they directly relate to the effective duration of ``memory'' in the neuron, and we hypothesise that optimizing these to characteristics will increase performance. 


To determine the effectiveness of the SNN approach, we can also turn the SNN network into a corresponding RNN network with RELU activation function by communicating to other neurons the membrane potential at every timestep rather than the occasional spike, replacing \eqref{eq2:adaptsn} with:
\begin{align}
S_{t} = RELU(u_t-\theta)
\label{eq:adaptsn-relu}
\end{align}


\paragraph{Backpropagation-Through-Time (BPTT)}
To train an SRNN, we apply Backpropagation-Through-Time (BPTT) \cite{werbos1990backpropagation,mozer1995focused}. With BPTT, the difference between the output predictions and output targets is propagated back from outputs to inputs, including past inputs, to optimize the weights and parameters by gradient descent. Conceptually, BPTT unrolls the network for all input timesteps. 

To compute the gradient through the discontinuous spiking mechanism, we apply the {\em surrogate gradient} approach as generalized in \cite{neftci2019surrogate} from earlier specific instances \cite{bohte2011error,bohte2000spikeprop}. To approximate the error-gradient through the discontinuous spike-generator of spiking neurons, the surrogate gradient approach substitutes this non-existing gradient with a derivative connecting the outgoing spike to the internal membrane potential. Multiple derivatives have been proposed \cite{neftci2019surrogate}; here, we use a Gaussian: $\hat{f_s}'(u_{t}) = \mathcal{N}(u_t|\theta,\sigma^2)$ where the mean of the distribution is $\theta$ -- the threshold, with standard deviation $\sigma$  -- the tolerance or variance -- used to scale the membrane potential for error-backpropagation. Unless stated otherwise, we set $\sigma$ to $0.5$. 


To define the loss-function that BPTT minimizes, we need to take into account the kind of task that is being performed and the kind of label that we have: for a sequential classification task, we receive a sequence of inputs and can make a decision at the end of the sequence. In a streaming task, we need to generate an output at every time step $t$. The loss-function is further defined by the error-metric, for which we can define a number of different approaches for interpreting the behavior of the output neurons. 

\paragraph{Decoding SRNNs} \label{sec:decoding}
Decoding the output of an SNN directly relates to the interpretation of the spiking neuron's behavior. Both the membrane potential trace and spike history, either in spike-counts and/or spike-timing, of the output neurons can be used to represent the belief of each class. We define a number of such output decoding methods and their associated loss-function.

{\bf Spike-based classification.}
When a neuron emits a spike, this is caused by the membrane potential (the hidden state of the neuron) reaching  threshold. In a classification task, a simple classification method is to count the number of spikes in a certain time-window. While straightforward, this method has some shortcomings, that may result in misclassifications: (1) some output neurons may fire the same number of spikes; (2) the reset and refectory mechanism of the spiking neuron may reduce the firing rate of a strong stimulus; and (3) real-time readout from single neurons is not feasible. As an alternative, we can use direct measures of the output neurons available at each time-step. 

{\bf Direct measures.} The membrane potential of an output neuron can also be used for classification, as it represents a measure of output neuron's stimulus history. We define several methods to decode the results from the membrane potential history:
\begin{enumerate}
  \item Last-timestep membrane potential: we take the value of the output membrane potential at the last time step of a sample as the output. Using a softmax activation, we then scale the outputs similar to the softmax function in ANNs. 
  \item Max-over-time membrane potential as: we take as the value of the output neurons the maximum membrane potential reached during presentation of the sample. 
  \item The readout integrator: while the membrane potential can be interpreted as a moving average of a neuron's activation, the resets caused by spiking do not fit in this notion. We define a non-spiking readout layer where the membrane potential is computed without the neuron spiking and resetting. This avoids the effect of reset mechanism of the spike neuron on classification performance. The readout integrator is defined as $u_t = \alpha u_t + (1-\alpha)x_t$ where $u_t$ is the output membrane potential and $x_t$ is the input spike train. $\alpha = \exp(-\tau_m/dt)$ where $\tau_m$ is a trainable time constant. We use the average value over time for the non-spiking readout neurons. 
\end{enumerate}
Variants of all three approaches were previously used in \cite{cramer2019heidelberg}; for streaming tasks, classifications are needed for every time-step, and when using only single neurons to represent outputs, we can only use the direct measurements.

To train the network, as in \cite{cramer2019heidelberg}, we use the cross-entropy function as the error function. In a streaming task, the readout membrane potentials are used as output and compared to the corresponding targets at each timestep. In the classification tasks, the output after reading the whole sequence is compared with the correct label of the sequence. Note that in \cite{cramer2019heidelberg}, cross-entropy is computed either for the max-over-time and last-time-step decoding. 

We implemented the various SRNNs in PyTorch, where 
the use of surrogate gradients allowed us to apply BPTT to minimize the loss efficiently, and also to leverage standard deep learning optimizations, including the training of spiking neuron parameters. 

%% file: 5_results.tex

\section{Experiments}

We apply the approach outlined above to a number of sequential classification and streaming classification tasks:  waveform classification in ECG signals in the QTDB dataset, the sequential and permuted sequential MNIST problem (S-MNIST, PS-MNIST), and the Spoken Heidelberg Digits (SHD) dataset.

{\bf Encoding and decoding.} SNNs as an event-driven neural network heavily rely on an encoding mechanism to convert external measures into spike-trains that feed into the network. Several approaches have been used to convert static or continuously changing values into spike-trains. In DVS-sensors \cite{lichtsteiner2008128}, a level-crossing scheme is used to encode a time-continuous signal into spikes; more generically, rate-based Poisson population encoding has been used \cite{OConnor2013-un,ruckauer2019closing}. 
Emperically, we found that different decoding schemes were best for different tasks: for the ECG task, decoding directly at every timestep from the membrane potential worked best, for the S-MNIST and PS-MNIST, it was spike-counting and for the SHD task the average readout integrator was most effective.

\paragraph{ECG} The analysis of electrocardiograms (ECGs) is widely used for monitoring the functionality of the cardiovascular system. As a kind of time-series data, ECG signals can be used to detect and recognize different waveforms and morphologies in heart-function. For a human, the recognition task is time-consuming and relies heavily on experience. 

{\bf Waveform classification.} In an ECG signal, there are three meaningful parts of cardiac period including $P$-wave, $T$-wave and the $QRS$-complex \cite{hurst1998naming}. In detail, the $QRS$ part consists of a $Q$-wave, an $R$-peak and an $S$-wave. In a monitoring task, we aim to continuously recognize the present type of wave. The streaming task is thus to character-wise recognize all six patterns of the ECG wave -- $P$, $PQ$, $QR$, $RS$, $ST$, and $TP$. An example of an ECG signal and the relative distribution of the 6 labels is shown in  Fig.~\ref{fig:dataset_ecg}.

\begin{table}[t]
  \small
\centering
{\setlength{\tabcolsep}{0.5em}
\bgroup
\def\arraystretch{0.98}
  \begin{tabular}{|l|l|r|r|r|}

  \hline
   Task                       & Method                    &  Network    & Acc.   & Fr \\ \hline  
    \multirow{7}{*}{ \makecell{ECG\\QTDB}}& Adaptive SRNN        & 46 &  \bf  84.4\% &.32\\ \cline{2-5}
                              & LIF SRNN       & 46      &   73.7\%  &.31\\ \cline{2-5}
                              & \em RELU SRNN        &46 &  {\bf \em  86.4\%}& \\ \cline{2-4}   
                              & \em LSTM                       &46 & \em  78.9\%& \\ \cline{2-4}
                              & \em GRU                        &46 & \em  77.3\% & \\ \cline{2-4}
                              & \em Vanilla RNN           &46 &  \em 74.8\%  &\\ \cline{2-4}
                              & \em Bid-LSTM$_{290}$   &290 &   \em  80.76\% & \\ \hline
\multirow{8}{*}{S-MNIST}  & Adaptive SRNN             &40(I)+256+128&  \bf 97.82\% &.077\\  \cline{2-5}  
                              & Adaptive SRNN   &80(I)+120+100 &  97.2\% &.075\\ \cline{2-5}  
                              & LIF SRNN                  &40(I)+256+128&  10$^{**}$\%& \\ \cline{2-4} 
                              & LSNN (L2L) \tiny \cite{bellec2018long}&80(I)+120+100&  93.7\% &\\ \cline{2-4}
                              & LSNN (L2L+DeepR) \tiny \cite{bellec2018long}      &80(I)+120+100&  96.4\% &\\ \cline{2-4}  
                              & \emph{RELU SRNN}       &  40(I)+256+128 & 98.99\% &\\ \cline{2-4} 
                              & \emph{Dense IndRNN \tiny \cite{li2019deep}}&   &  \bf \em 99.48\% &\\ \cline{2-4}
                              & \emph{LSTM    \tiny \cite{arjovsky2016unitary}}   & 128 & 98.2\% &\\ \cline{1-5}
                              
\multirow{7}{*}{PS-MNIST} 
                              & Adaptive SRNN       &40(I)+256+128&  91.0\% &.102\\ \cline{2-5}
                              & LIF SRNN     &80(I)+512+50&  10$^{**}$\% &\\ \cline{2-4}
                               &\emph{RELU SRNN } &40(I)+256+128&  93.47\% &\\ \cline{2-4}
                              & \emph{Dense IndRNN  \tiny \cite{li2019deep}} & &97.2\%&\\ \cline{2-4}
                              & \emph{LSTM \tiny \cite{arjovsky2016unitary}}     &  128  &  88\%   & \\ \hline
    \multirow{9}{*}{SHD}  & Adaptive SRNN (4ms)                          &  128 & 79.4\% &.071\\ \cline{2-5}
                              & Adaptive SRNN (4ms)                          &  256 & 81.71\% &.049\\ \cline{2-5}
                              & Adaptive SRNN (4ms)                          &  128+128 & \bf 84.4\% &.103 \\ \cline{2-5}
                              & LIF SRNN (4ms)                               &  256 & 78.93\% &.021\\ \cline{2-5}
                              & LIF RSNN  \tiny \cite{cramer2019heidelberg}        &  128*3 &71.4\%  &\\ \cline{2-4} 
                              & \em RELU SRNN (4ms)                           & 128+128 & \bf 88.93\%& \\ \cline{2-4}
                              & \em LSTM  \tiny \cite{cramer2019heidelberg}$^+$         &  128   &   85.7\% &\\  \cline{2-4} 
                              & \em CNN$^+$ \tiny \cite{cramer2019heidelberg}          & 1,014,036& \bf 92.4\%  &\\ \cline{2-4}
                              \hline
\end{tabular}}
\egroup
  \caption{ECG, S-MNIST, PS-MNIST and SHD results. (I) denotes the number of population input coding. $^{**}$: the network did not learn. Bid-LSTM is a Bi-directional LSTM neural network. Model$^+$ is using the binned count as encoding method. Italic methods are ANN networks. Fr denotes the average sparsity in the SNN (spikes per neuron per time-step).\vspace{-0.5cm} }
  \label{tab:smnistsota}
\end{table}

{\bf QTDB} is the one of the most widely used ECG datasets for wave segmentation, where the data is well labeled in the temporal dimension. Each sample has two channels -- 'a' and 'b', and this provide additional spatial information. The original data consist of float values for each timestep; to convert this signal into an event-based one, we applied {\em level-crossing encoding} on the ECG signal to convert the continuous values into spike-trains. Level-crossing is applied to the normalized ECG signal by converting each channel signal into two spike channels representing increasing and decreasing event respectively. A spike is generated when the amplitude increase or decrease is larger than a threshold -- here, we use 0.3 as a threshold. The result is a compression of the ECG signal by about $75\%$. 

We apply several RNNs and SRNNs for ECG waveform classification, see Table \ref{tab:smnistsota}. We find that the SRNNs with adaptive spiking neurons achieved the best performance of $84.4\%$ with the smallest size neural network of 46 neurons (36 hidden, 4 input and 4 output neurons). A same-sized SRNN comprised of LIF neurons achieved only $73.7\%$. The LSTM and vanilla RNN with the same network structure achieved  $78.9\%$, $77.3\%$ and $74.8\%$ accuracy respectively; a birectional-LSTM with 290 units achieved $80.76\%$ .
The best performance ($86.4\%$) was obtained by turning the adaptive SRNN into an ANN, the RELU SRNN. 

The accuracy results are presented by evaluating the input that had been fed in at every time step, which is sampled at 250 Hz. No delay between input and output evaluation have been taken into account. In contrast to the spikes input generated by the level-crossing encoding that the SRNN receives as input, the LSTM and RNN networks receive floating point values at their inputs. These values represent the ADC sample values (12 bits precision).

\paragraph{S-MNIST and PS-MNIST} The MNIST dataset is the seminal computer vision classification task. The Sequential MNIST (S-MNIST) benchmark and Permuted MNIST (PS-MNIST) benchmark were introduced as corresponding problems for sequential data processing \cite{le2015simple}, presenting each pixel in an MNIST image pixel by pixel, resulting in a sequence of length $28\times28=784$. 


\begin{figure}
    \centering
    \begin{minipage}{.45\linewidth}
            \begin{subfigure}[t]{.9\linewidth}
                \includegraphics[width=\textwidth]{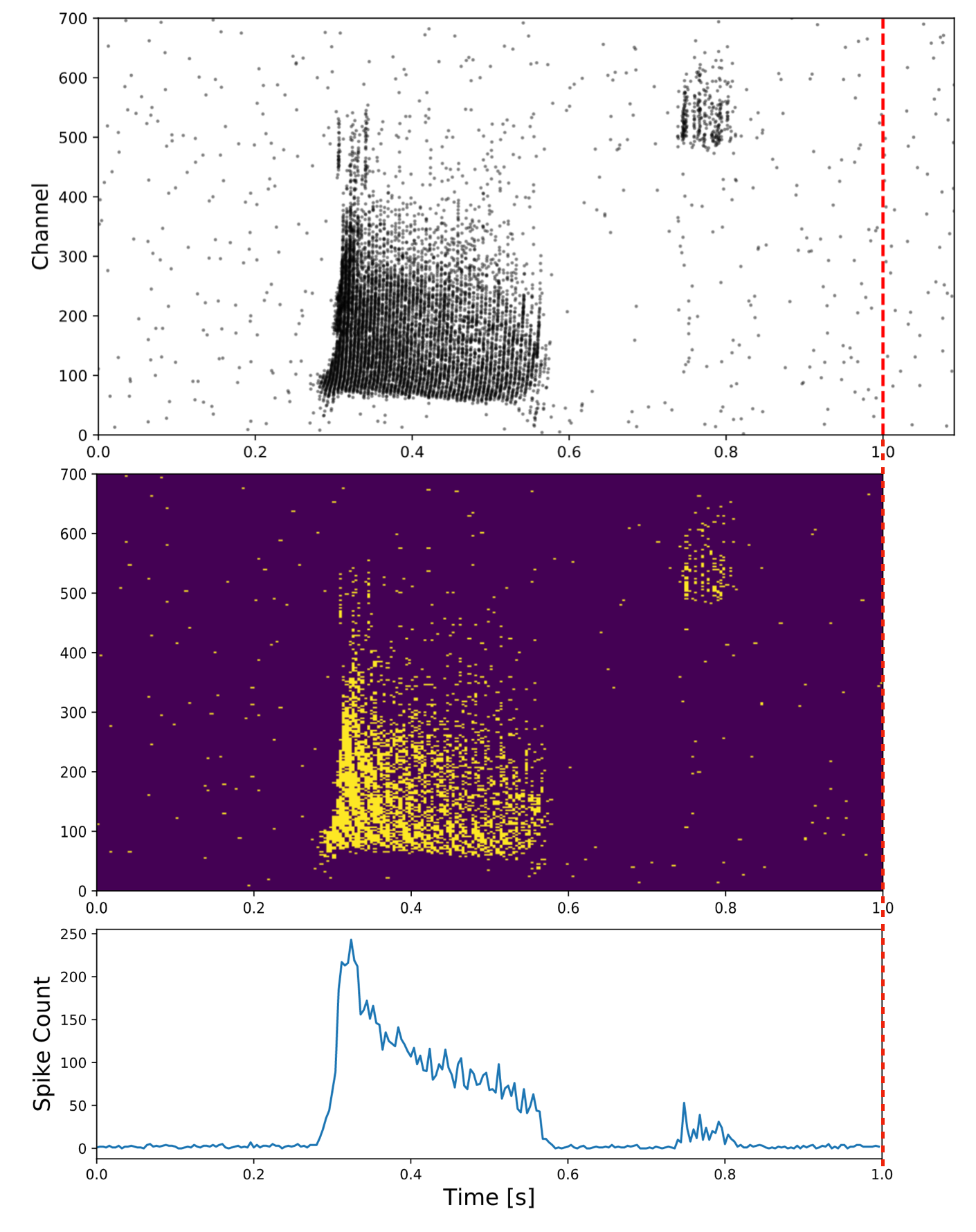}
                \caption{An example from preprocessed and padded SHD dataset. \textbf{Top}: a sample from the SHD dataset, \textbf{Middle}: corresponding processed input spike trains, \textbf{Bottom}: the spike count summed over all 700 input channels.}
                \label{fig:SHD_sample}
            \end{subfigure}
        \end{minipage}
    \begin{minipage}{.45\linewidth}
        \begin{subfigure}[t]{.9\linewidth}
            \includegraphics[width=\textwidth]{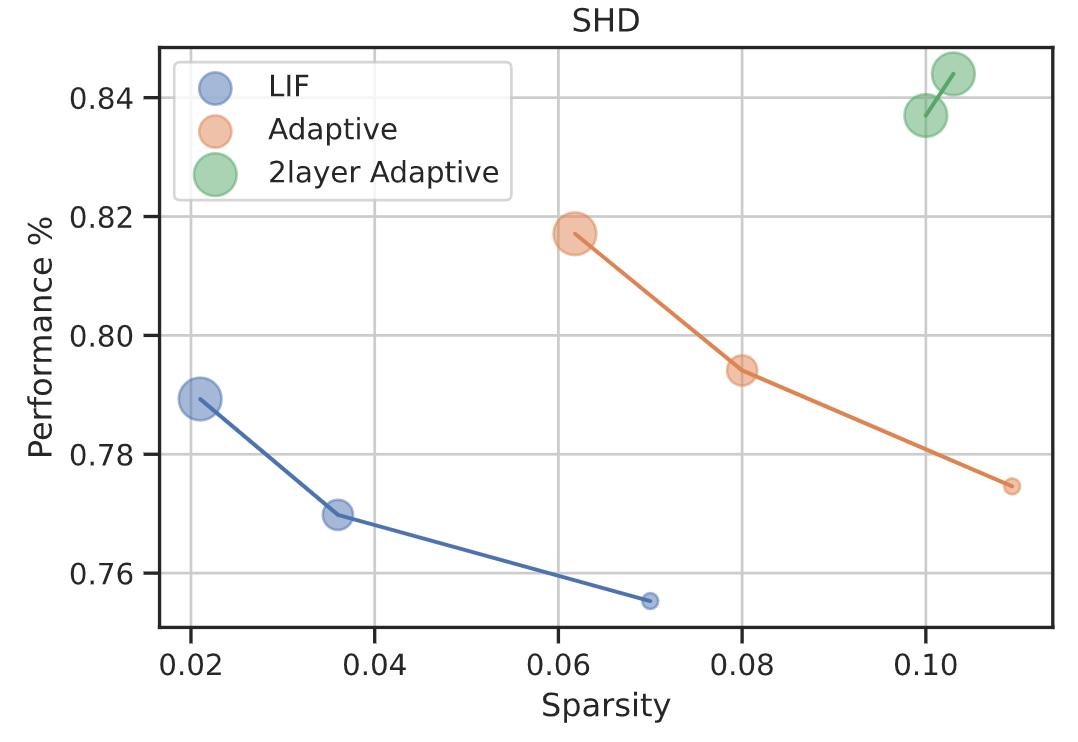}
            \caption{SRNN performance versus sparsity for the SHD task. The node size indicates network size.}
            \label{fig:SHD-acc-fr}
        \end{subfigure} \\
        \begin{subfigure}[b]{.9\linewidth}
            \includegraphics[width=\textwidth]{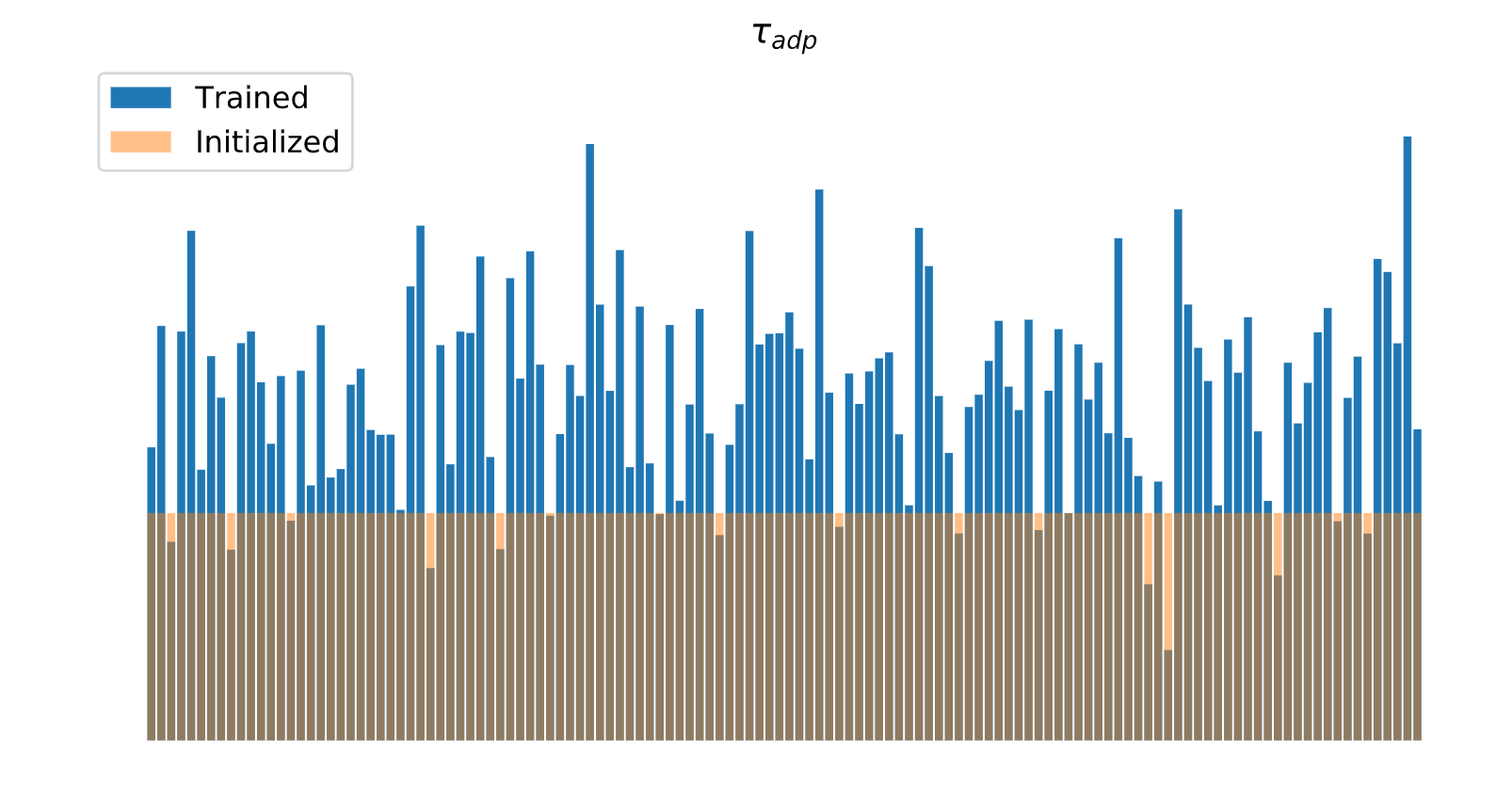}
            \caption{Histogram of $\tau_{adp}$ values at network initialization (yellow) and after training (blue) on the SHD task.}
            \label{fittg:tau_adp_hist}
        \end{subfigure} 
    \end{minipage}
    \caption{(a)SHD example;(b)SRNN performance on SHD dataset;(c)Evolving of $\tau_{adp}$.\vspace{-0.5cm}}
\end{figure}

For {\bf S-MNIST} the state of the art accuracy is $99.48\%$ obtained with the Dense IndRNN \cite{li2019deep}; the best reported performance of an LSTM is $98.2\%$ \cite{arjovsky2016unitary}. For these RNNs, the analogue grey pixel value is directly fed as input into the network. With SNNs, Bellec {\em et al.} \cite{bellec2018long} obtained $93.7\%$ on the S-MNIST task using eProp and population Poisson encoding to encode the grey pixel values, and $96.4\%$ when additionally using a learning-2-learn meta-learning framework. As can be seen in Table \ref{tab:smnistsota}, our adaptive SRNNs using two recurrent layers obtained $97.82\%$ with the same encoding scheme; with the same network layout and size as  \cite{bellec2018long} the SRNN achieved $97.2\%$. When using LIF neurons in various SRNN architectures, the networks failed to learn; turning the adaptive SRNN into an ANN again however increased performance to $98.99\%$, approaching the Dense IndRNN accuracy.

{\bf PS-MNIST} is a harder problem than S-MNIST, as first a permutation is applied to all images before sequentially reading the image pixel-by-pixel \cite{le2015simple}. The permutation strongly distorts the temporal pattern in the input sequence, making the task more difficult than S-MNIST. The Dense IndRNN  \cite{li2019deep} here obtained $97.2\%$ accuracy; the LSTM \cite{arjovsky2016unitary} achieved only $88\%$. We are not aware of any SNNs benchmark data on this task; our adaptive SRNN achieved $91.0\%$ accuracy on the test dataset; the LIF SRNN again failed to learn while the RELU SRNN obtained $93.47\%$.

\paragraph{SHD}
The Spoken Heidelberg Digits spiking dataset was developed specifically for benchmarking spiking neural networks \cite{cramer2019heidelberg}. It was created based on the Heidelberg Digits (HD) audio dataset which comprises of 20 classes of spoken digits from $zero$ to $nine$ in English and German, spoken by 12 individuals. For training and evaluation, the dataset (10420 samples) is split into a training set (8156 samples) and test set (2264 samples). An LSTM with 128 units  achieved $85.7\%$ \cite{cramer2019heidelberg}, where the continuous time stream is binned into 10ms segments and the spike-count in each bin was used as input for the LSTM. For comparison, treating each sample as an image to train a deep CNN with over 1 million neurons achieves $92.4\%$ \cite{cramer2019heidelberg}. Using a three layer spiking recurrent network comprised of LIF neurons with 128 neurons in each layer, \cite{cramer2019heidelberg} obtained $71.4\%$.

To apply our SRNNs, we converted all audio samples into 250-by-700 binary matrices. For this, all samples were fit within a 1 second window; shorter samples were padded with zeros and longer samples were cut by removing the tail (the latter applied to only 20 samples, with the longest sampling being 1.17s; visual inspection showed no significant data in the tail -- an example is shown in Fig~\ref{fig:SHD_sample}). Spikes were then binned in time bins both of size 10ms and 4ms; for the SRNNs, the presence or non-presence of any spikes in the time-bin is noted as a single binary event; for the LSTMs, the spike-count in a bin is used as the (binary) input value. During training, a $10\%$ subset of the original training dataset was used for validation, with Adam \cite{kingma2014adam} as the default optimizer. The initial learning rate is set to 0.01 with a $50\%$ decay at epoch 10, 50, 120 and 200. For the non-spiking RELU SRNN, 50 training epochs sufficed. We trained SRNNs both with a single recurrent layer and with two recurrent layers, with LIF or adaptive spiking neurons. The membrane potential of each SRNN neuron was set to a random number between 0 and 1 at the start of each sample. 

For an adaptive SRNN with two layers of 128 adaptive spiking neurons, trained on the 4ms binned data, we obtained $84.4\%$ accuracy, approaching the 128 unit LSTM in \cite{cramer2019heidelberg}. 
An adaptive SRNN with a single layer of 256 spiking neurons achieved $81.71\%$, demonstrating the utility of having multiple recurrent layers in an SRNN. Similarly, a LIF SRNN with a single recurrent layer of 256 neurons achieves only $78.93\%$. The non-spiking RELU SRNN substantially outperforms the spiking SRNN, obtaining an accuracy of $88.93\%$.

\paragraph{Sparsity}
In Table \ref{tab:smnistsota}, we also note the sparsity (Fr) of the trained SRNNs, where sparsity is defined as the percentage of active neurons at each step.  We find that for the ECG task, neurons fire on average once every 3 timesteps (FR=0.32), where this relatively low sparsity is likely caused by the need to read out class labels at every timestep. For the MNIST tasks, sparsity is much higher, varying between 0.07 and 0.1, and for the SHD task, sparsity varies between 0.02 and 0.13, mostly as a function of accuracy. 

For SHD, we investigated the relationship between network performance and sparsity for different size networks in more detail. In Fig \ref{fig:SHD-acc-fr}, we see that the network performance can be increased and sparsity improved at the same time by increasing the size of the network. We also see that the performance advantage of adaptive neurons compared to LIF neuron comes at the expense of sparsity. 



\begin{figure}[t!p]
    \centering
    \includegraphics[width=1.\columnwidth]{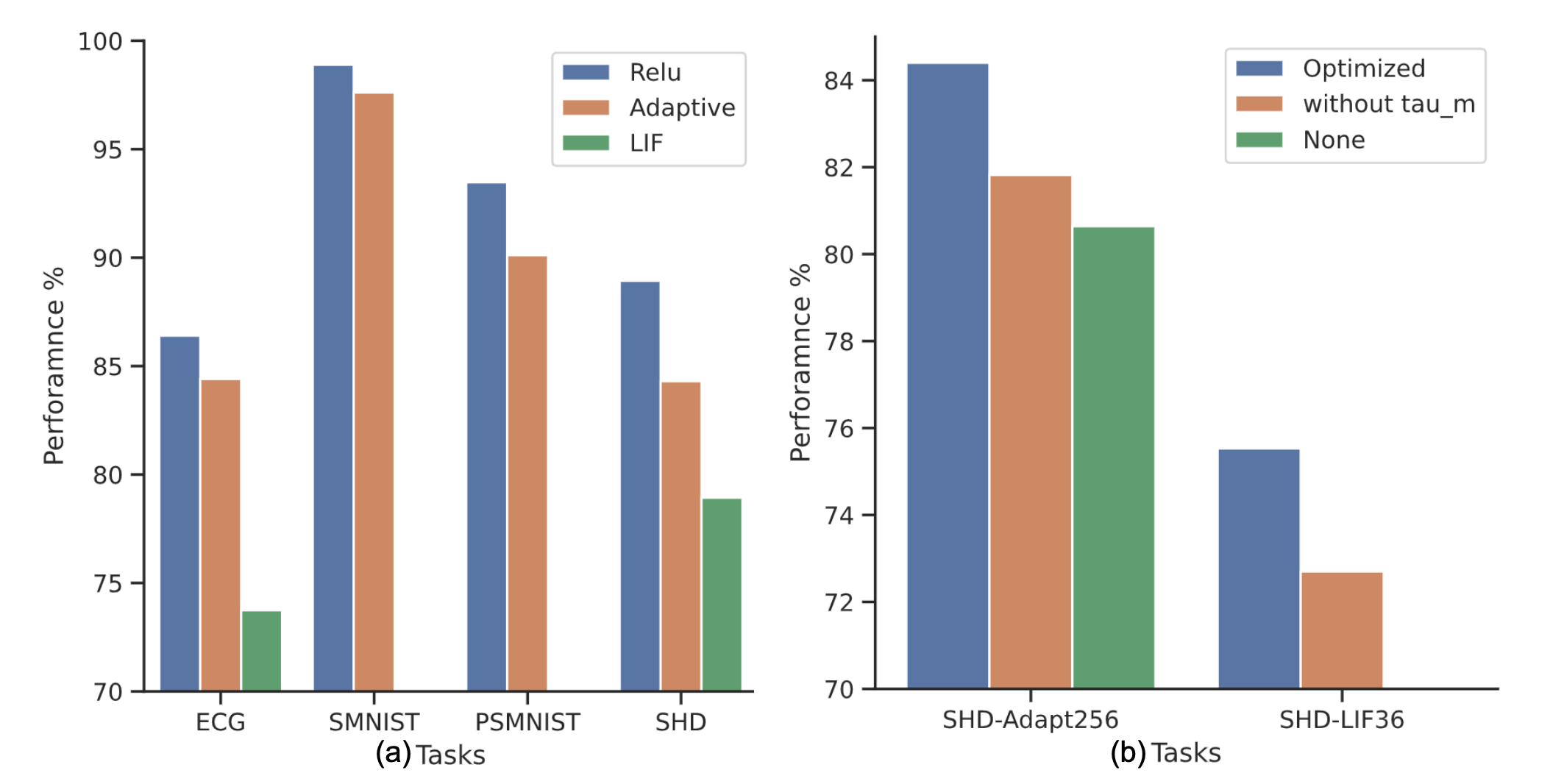}
    \vspace{-0.6cm}
    \caption{(a) Effect of neuron types on performance for each task; (b) Successive ablation (non-training) of training time-constants in the adaptive and LIF neurons on the SHD task.}
    \vspace{-0.3cm}
    \label{fig:neuron_acc}
\end{figure}

\paragraph{Complexity of Spiking Neurons}
In general, we find that using adaptive spiking neurons with time-constants adjusted during training substantially outperform LIF neurons, as illustrated in Fig \ref{fig:neuron_acc}(a). As can be seen in Fig \ref{fig:neuron_acc}(b), training these time-constants substantially improved performance, illustrated for the SHD dataset, as successive ablation of training these parameters reduces performance. As illustrated in Fig \ref{fittg:tau_adp_hist}, training also substantially modifies these parameters: shown is both the initial histogram of $\tau_{adp}$ in the SHD task and the histogram after training. 


%% file: Energy.tex
\section{Efficiency}
A hotly debated topic is whether or not SNNs can achieve a meaningful power-reduction compared to ANNs \cite{panda2019towards}. Here we derive theoretical energy values based on power numbers at the register transfer logic (RTL) level for 45nm CMOS technology from \cite{panda2019towards}.

\begin{table}[b]
\small
\centering
\begin{tabular}{|l|l|l|}
\hline
Network     & Recurrent    & Energy/layer                \\ \hline
LIF         &Yes    &     $(mn+nn)E_{AC}Fr$         \\ \hline
Adaptive    &Yes    &     \makecell[l]{$(mn+nn+2n)E_{AC}Fr$\\$+2nE_{MAC}$}        \\ \hline
LIF         &No    &      $(mn)E_{AC}Fr$           \\ \hline
Adaptive    &No    &      \makecell[l]{$(mn+2n)E_{AC}Fr$\\$+2nE_{MAC}$}          \\ \hline
\hline
RNN         &Yes    &     $(mn+nn)E_{MAC}$         \\ \hline 
Adaptive$^*$&Yes    &     $(mn+nn+4n)E_{MAC}$        \\ \hline
LSTM        &Yes    &     $(4mn+4nn+3n)E_{MAC}$         \\ \hline
Bir-LSTM    &Yes    &     $2(4mn+4nn+3n)E_{MAC}$       \\ \hline
\end{tabular}
\caption{Energy consumption per layer for various neurons. The network layer $l$ with input size is $m$ and output size is $n$. $E_{AC}$ is the energy cost per AC, $E_{MAC}$ the cost per MAC. Adaptive$^*$ is the non spiking adaptive neuron. }
\vspace{-0.5cm}
\label{Tab:flops-layer-table}
\end{table}

\begin{figure*}[t!p]
    \centering
    \includegraphics[width=\textwidth]{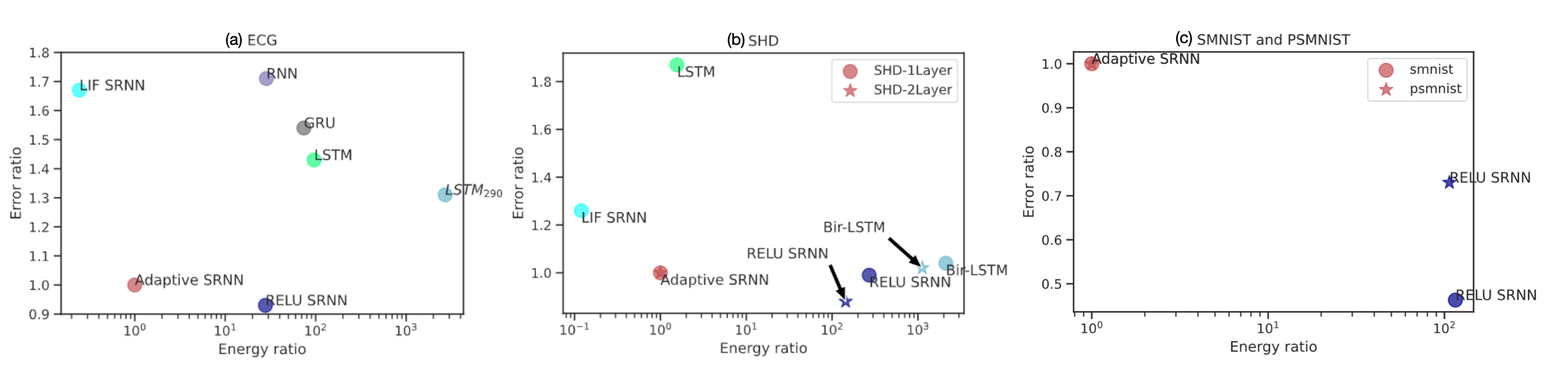}
    \caption{Scatter plot of energy and error ratio on each task. The Adaptive SRNN model is used as unit baseline performance for both energy and error ratio.}
    \label{fig:energy-error}
\end{figure*}

We calculate the theoretical energy consumption of a recurrent network by counting the required operations per timestep. We count both multiply-and-accumulate operations (MACs) and accumulate (AC) operations. \cite{panda2019towards}. A standard artificial neuron requires a MAC for each input; in contrast, a spiking neuron only requires an accumulate (AC) for each input spike, while it's internal state dynamics require some MACs. 

In a network, we thus need to consider the number fan-in connection into a neuron, the number of neurons in a layer, and the cost internal calculations. For example, consider a recurrent operation at layer $l$ that is defined as $y_{l,t} = f(W \cdot x_{t} + W_{rec} \cdot y_{l,t-1})+b$ with input size $m$ and output size $n$: this requires two multiply operations and one accumulate operation. The energy required for the RNN is then computed as $Energy_{rnn} = (mn+nn)E_{MAC}$, for every timestep. In the SNN however, the sparse spiking activity of the network (the average Firing Rate $Fr$) needs to be considered: $Energy_{srnn} = (mn+nn)E_{AC}Fr,$ with $Fr<<1$ in SRNNs with sparse activity.

We compute the theoretical energy cost of a recurrent network $RNN$ as the sum over all $L$ layers and all $T$ time steps, $E_{rnn} = \sum_{t\in T}\sum_{l\in L} Energy_{l,t}$: we computed the MACs/ACs and energy use for various recurrent networks in Table \ref{Tab:flops-layer-table}. For the network architectures used in this study, we then calculate the actual relative energy cost in Table~\ref{tab:sota-ratio}, and we plot the Accuracy vs Energy ratio for the various networks in Fig~\ref{fig:energy-error}. In Fig~\ref{fig:energy-error}, we see that our SRNN solutions lie on the Pareto front of energy efficient and effective networks, with the spiking adaptive SRNN achieving close to the RELU SRNN performance while theoretically being 28--243x more energy efficient. Compared to more classical RNNs  on the more complex SHD and (P)S-MNIST tasks, we calculate the SRNNs to be $>$100x more energy efficient as in these larger networks both the fan-in factor and sparsity increases.



\begin{table*}[t]
\small
  \centering
 {\setlength{\tabcolsep}{0.6em}
\bgroup
\def\arraystretch{0.985} 
  \begin{tabular}{|l|l|r|r|r|r|r|}
  \hline
   Task                       & Method  & Accuracy &Energy$_{nn}$/s &\makecell{Energy \\ratio} & \makecell{Error\\ratio} & Efficiency\\ \hline
 \multirow{8}{*}{\bf ECG-qtdb}& Adapt. SRNN  &\bf \em 84.4\% &208           &1x           &1x     & 1x\\ \cline{2-7}                              &LIF SRNN &73.7                 &51           &.246x      &1.69x   &.416x\\ \cline{2-7} 
                              & RELU SRNN          &  86.4\%  &5,837         &28x        &.872x   &24.4x\\ \cline{2-7}   
                              & LSTM           &  78.9\%  &19,469            &93.5x        &1.35x  &126.2x\\ \cline{2-7}
                              & Vanilla RNN  &  74.8\%  &5,299                &25.5x        &1.62x  &41.3x\\ \cline{2-7}
                              & Bid. LSTM$_{290}$ &  80.76\%& 567,424         &2726x   &1.23x  &3353.3x\\ \cline{1-7}
 \multirow{6}{*}{\bf SHD 256}& Adapt. SRNN &\bf \em 81.71\%  & 3,249          &1x           &1x     & 1x\\ \cline{2-7}
                              & LIF SRNN            &  78.93\% &589             &.181x      &1.15x &.21x\\ \cline{2-7} 
                              & RELU SRNN           &  83.9\%   &802,880           &247x        &.88x &217x\\ \cline{2-7}   
                              & LSTM        &  69.71\%  &3,135,144     &965x      &1.66x  &1602x\\ \cline{2-7}
                              & Bid. LSTM&  83.17\%  &6,270,222      & 1930x        &.92x       &1776x\\ \cline{1-7} 
 \multirow{3}{*}{\bf SHD-2layer}& Adapt. SRNN &\bf \em 84.4\%  &3120.6           &1x           &1x     & 1x\\  \cline{2-7}
                              & RELU SRNN       &  88.93\%   &45,539          &146x         &.71x      &104x\\ \cline{2-7}   
                              & Bid. LSTM &  87.23\%   &3,557,156        &1140x       &.819x    &935x\\ \cline{1-7} 
 \multirow{2}{*}{\bf S-MNIST}  & Adapt. SRNN &\bf \em 97.82\%  &3410            &1x           &1x     & 1x\\  \cline{2-7}
                              & RELU SRNN         &  98.99\%   &  408,908       &120x        &.463x     &55.6x\\  \cline{1-7} 
 \multirow{2}{*}{\bf PS-MNIST}  & Adapt. SRNN &\bf \em 91.0\%  & 3791           &1x           &1x     & 1x\\  \cline{2-7}
                              & RELU SRNN      &  93.47\%   &  408,908       &108x        &.73x     &78.8x\\  \cline{1-7} 
                              \hline
     \end{tabular}}
     \egroup
  \caption{Model performance and energy consumption. For comparison, the energy consumption and error rate of the Adaptive SRNN was set to 1x. The energy/error ratio is computed as the ratio of the energy/error relative to the adaptive SRNN. Efficiency is defined as the product of energy and error ratio. For energy consumption measures $E_{MAC}$ and $E_{AC}$, we use \cite{panda2019towards}: a 32-bit MAC Int requires 3.2pJ and a 32-bit AC Int require 0.1pJ.}
  \label{tab:sota-ratio}
\end{table*}

%% file: 6_Conclusions.tex
\section{Conclusions}
We demonstrated how competitive recurrent spiking neural networks can be trained using backpropagation-through-time (BPTT) and surrogate gradients on classical and novel time-continuous tasks, achieving novel state-of-the-art for spiking neural networks and approaching or exceeding state-of-the-art for classical RNNs on these tasks. Calculating the theoretical energy cost, we find that our spiking SRNNs are up to 243X more efficient than the slightly better performing analog RELU SRNN, and up to 1900x times more efficient than similarly performing classical RNNs like LSTMs. 

We showed that using more complex adaptive spiking neurons was key to achieving these results, in particular by also training the individual time-constants of these spiking neurons, also using BPTT. Having two time-constants, the adaptive spiking neuron effectively maintains a multiple-timescale memory. We hypothesise that this approach is so effective because it allows the memory in the network to be adapted to the temporal dynamics of the task. Surprisingly, converting the SRNN to a non-spiking RELU RNN consistently increased performance, suggesting that the nested hierarchical recurrent network architecture is particularly effective. 

Training these complex SRNNs including the various parameters was only feasible because, by using surrogate gradients, we were able to use a mature and advanced deep learning framework (PyTorch) and benefit from the automated differentiation to also train spiking neuron parameters\footnote{code is available at \url{https://github.com/byin-cwi/SRNN-ICONs2020}}. We believe this approach opens up new opportunities for improving and scaling SNNs further.  

\section{Acknowledgements}
BY is funded by the NWO-TTW Programme “Efficient Deep Learning” (EDL) P16-25.
The authors gratefully acknowledges the support from the organizers of the Capo Caccia Neuromorphic Cognition 2019 workshop and Neurotech CSA, as well as Jibin Wu and Saray Soldado Magraner for helpful discussions.

%% file: main.bbl
\begin{thebibliography}{10}

\bibitem{Roy2019-rv}
Kaushik Roy, Akhilesh Jaiswal, and Priyadarshini Panda.
\newblock Towards spike-based machine intelligence with neuromorphic computing.
\newblock {\em Nature}, 575(7784):607--617, November 2019.

\bibitem{courbariaux2015binaryconnect}
Matthieu Courbariaux, Yoshua Bengio, and Jean-Pierre David.
\newblock Binaryconnect: Training deep neural networks with binary weights
  during propagations.
\newblock In {\em Advances in neural information processing systems}, pages
  3123--3131, 2015.

\bibitem{gong2019differentiable}
Ruihao Gong, Xianglong Liu, Shenghu Jiang, Tianxiang Li, Peng Hu, Jiazhen Lin,
  Fengwei Yu, and Junjie Yan.
\newblock Differentiable soft quantization: Bridging full-precision and low-bit
  neural networks.
\newblock In {\em Proceedings of the IEEE International Conference on Computer
  Vision}, pages 4852--4861, 2019.

\bibitem{rastegari2016xnor}
Mohammad Rastegari, Vicente Ordonez, Joseph Redmon, and Ali Farhadi.
\newblock Xnor-net: Imagenet classification using binary convolutional neural
  networks.
\newblock In {\em European conference on computer vision}, pages 525--542.
  Springer, 2016.

\bibitem{darabi2018bnn+}
Sajad Darabi, Mouloud Belbahri, Matthieu Courbariaux, and Vahid~Partovi Nia.
\newblock Bnn+: Improved binary network training.
\newblock {\em arXiv preprint arXiv:1812.11800}, 2018.

\bibitem{tan2019efficientnet}
Mingxing Tan and Quoc~V Le.
\newblock Efficientnet: Rethinking model scaling for convolutional neural
  networks.
\newblock {\em arXiv preprint arXiv:1905.11946}, 2019.

\bibitem{yang2017designing}
Tien-Ju Yang, Yu-Hsin Chen, and Vivienne Sze.
\newblock Designing energy-efficient convolutional neural networks using
  energy-aware pruning.
\newblock In {\em Proceedings of the IEEE Conference on Computer Vision and
  Pattern Recognition}, pages 5687--5695, 2017.

\bibitem{maass1997networks}
Wolfgang Maass.
\newblock Networks of spiking neurons: the third generation of neural network
  models.
\newblock {\em Neural networks}, 10(9):1659--1671, 1997.

\bibitem{davies2018loihi}
Mike Davies, Narayan Srinivasa, Tsung-Han Lin, Gautham Chinya, Yongqiang Cao,
  Sri~Harsha Choday, Georgios Dimou, Prasad Joshi, Nabil Imam, Shweta Jain,
  et~al.
\newblock Loihi: A neuromorphic manycore processor with on-chip learning.
\newblock {\em IEEE Micro}, 38(1):82--99, 2018.

\bibitem{panda2019towards}
Priyadarshini Panda, Aparna Aketi, and Kaushik Roy.
\newblock Towards scalable, efficient and accurate deep spiking neural networks
  with backward residual connections, stochastic softmax and hybridization.
\newblock {\em arXiv preprint arXiv:1910.13931}, 2019.

\bibitem{bohte2011error}
Sander~M Bohte.
\newblock Error-backpropagation in networks of fractionally predictive spiking
  neurons.
\newblock In {\em International Conference on Artificial Neural Networks},
  pages 60--68. Springer, 2011.

\bibitem{neftci2019surrogate}
Emre~O Neftci, Hesham Mostafa, and Friedemann Zenke.
\newblock Surrogate gradient learning in spiking neural networks.
\newblock {\em arXiv preprint arXiv:1901.09948}, 2019.

\bibitem{bohte2000spikeprop}
Sander~M Bohte, Joost~N Kok, and Johannes~A La~Poutr{\'e}.
\newblock Spikeprop: backpropagation for networks of spiking neurons.
\newblock In {\em ESANN}, volume~48, pages 17--37, 2000.

\bibitem{werbos1990backpropagation}
Paul~J Werbos.
\newblock Backpropagation through time: what it does and how to do it.
\newblock {\em Proceedings of the IEEE}, 78(10):1550--1560, 1990.

\bibitem{mozer1995focused}
Michael~C Mozer.
\newblock A focused backpropagation algorithm for temporal.
\newblock {\em Backpropagation: Theory, architectures, and applications}, 137,
  1995.

\bibitem{hochreiter1997long}
Sepp Hochreiter and J{\"u}rgen Schmidhuber.
\newblock Long short-term memory.
\newblock {\em Neural computation}, 9(8):1735--1780, 1997.

\bibitem{graves2013generating}
Alex Graves.
\newblock Generating sequences with recurrent neural networks.
\newblock {\em arXiv preprint arXiv:1308.0850}, 2013.

\bibitem{li2019deep}
Shuai Li, Wanqing Li, Chris Cook, Yanbo Gao, and Ce~Zhu.
\newblock Deep independently recurrent neural network (indrnn).
\newblock {\em arXiv preprint arXiv:1910.06251}, 2019.

\bibitem{oord2016wavenet}
Aaron van~den Oord, Sander Dieleman, Heiga Zen, Karen Simonyan, Oriol Vinyals,
  Alex Graves, Nal Kalchbrenner, Andrew Senior, and Koray Kavukcuoglu.
\newblock Wavenet: A generative model for raw audio.
\newblock {\em arXiv preprint arXiv:1609.03499}, 2016.

\bibitem{benidis2020neural}
Konstantinos Benidis, Syama~Sundar Rangapuram, Valentin Flunkert, Bernie Wang,
  Danielle Maddix, Caner Turkmen, Jan Gasthaus, Michael Bohlke-Schneider, David
  Salinas, Lorenzo Stella, et~al.
\newblock Neural forecasting: Introduction and literature overview.
\newblock {\em arXiv preprint arXiv:2004.10240}, 2020.

\bibitem{maass2001pulsed}
Wolfgang Maass and Christopher~M Bishop.
\newblock {\em Pulsed neural networks}.
\newblock MIT press, 2001.

\bibitem{attwell2001energy}
David Attwell and Simon~B Laughlin.
\newblock An energy budget for signaling in the grey matter of the brain.
\newblock {\em Journal of Cerebral Blood Flow \& Metabolism},
  21(10):1133--1145, 2001.

\bibitem{bellec2018long}
Guillaume Bellec, Darjan Salaj, Anand Subramoney, Robert Legenstein, and
  Wolfgang Maass.
\newblock Long short-term memory and learning-to-learn in networks of spiking
  neurons.
\newblock In {\em Advances in Neural Information Processing Systems}, pages
  787--797, 2018.

\bibitem{zenke2018superspike}
Friedemann Zenke and Surya Ganguli.
\newblock Superspike: Supervised learning in multilayer spiking neural
  networks.
\newblock {\em Neural computation}, 30(6):1514--1541, 2018.

\bibitem{Shrestha2018}
Sumit~Bam Shrestha and Garrick Orchard.
\newblock {SLAYER}: Spike layer error reassignment in time.
\newblock In S.~Bengio, H.~Wallach, H.~Larochelle, K.~Grauman, N.~Cesa-Bianchi,
  and R.~Garnett, editors, {\em Advances in Neural Information Processing
  Systems 31}, pages 1419--1428. Curran Associates, Inc., 2018.

\bibitem{huh2018gradient}
Dongsung Huh and Terrence~J Sejnowski.
\newblock Gradient descent for spiking neural networks.
\newblock In {\em Advances in Neural Information Processing Systems}, pages
  1433--1443, 2018.

\bibitem{ponulak2008analysis}
Filip Ponulak.
\newblock Analysis of the resume learning process for spiking neural networks.
\newblock {\em International Journal of Applied Mathematics and Computer
  Science}, 18(2):117--127, 2008.

\bibitem{zambrano2016fast}
Davide Zambrano and Sander~M Bohte.
\newblock Fast and efficient asynchronous neural computation with adapting
  spiking neural networks.
\newblock {\em arXiv preprint arXiv:1609.02053}, 2016.

\bibitem{tavanaei2019deep}
Amirhossein Tavanaei, Masoud Ghodrati, Saeed~Reza Kheradpisheh, Timothee
  Masquelier, and Anthony Maida.
\newblock Deep learning in spiking neural networks.
\newblock {\em Neural Networks}, 111:47--63, 2019.

\bibitem{Wang2020-uq}
Wei Wang, Shibo Zhou, Jingxi Li, Xiaohua Li, Junsong Yuan, and Zhanpeng Jin.
\newblock Temporal pulses driven spiking neural network for fast object
  recognition in autonomous driving.
\newblock {\em arXiv preprint arXiv:2001.09220}, 2020.

\bibitem{cramer2019heidelberg}
Benjamin Cramer, Yannik Stradmann, Johannes Schemmel, and Friedemann Zenke.
\newblock The heidelberg spiking datasets for the systematic evaluation of
  spiking neural networks.
\newblock {\em arXiv preprint arXiv:1910.07407}, 2019.

\bibitem{Gerstner2002-wd}
Wulfram Gerstner and Werner~M Kistler.
\newblock {\em Spiking Neuron Models: Single Neurons, Populations, Plasticity}.
\newblock Cambridge University Press, August 2002.

\bibitem{izhikevich2003simple}
Eugene~M Izhikevich.
\newblock Simple model of spiking neurons.
\newblock {\em IEEE Transactions on neural networks}, 14(6):1569--1572, 2003.

\bibitem{lichtsteiner2008128}
Patrick Lichtsteiner, Christoph Posch, and Tobi Delbruck.
\newblock A 128x128 120 db 15$\mu$s latency asynchronous temporal contrast
  vision sensor.
\newblock {\em IEEE journal of solid-state circuits}, 43(2):566--576, 2008.

\bibitem{OConnor2013-un}
Peter O'Connor, Daniel Neil, Shih-Chii Liu, Tobi Delbruck, and Michael
  Pfeiffer.
\newblock Real-time classification and sensor fusion with a spiking deep belief
  network.
\newblock {\em Front. Neurosci.}, 7:178, October 2013.

\bibitem{ruckauer2019closing}
Bodo R{\"u}ckauer, Nicolas K{\"a}nzig, Shih-Chii Liu, Tobi Delbruck, and Yulia
  Sandamirskaya.
\newblock Closing the accuracy gap in an event-based visual recognition task.
\newblock {\em arXiv preprint arXiv:1906.08859}, 2019.

\bibitem{hurst1998naming}
J~Willis Hurst.
\newblock Naming of the waves in the ecg, with a brief account of their
  genesis.
\newblock {\em Circulation}, 98(18):1937--1942, 1998.

\bibitem{arjovsky2016unitary}
Martin Arjovsky, Amar Shah, and Yoshua Bengio.
\newblock Unitary evolution recurrent neural networks.
\newblock In {\em International Conference on Machine Learning}, pages
  1120--1128, 2016.

\bibitem{le2015simple}
Quoc~V Le, Navdeep Jaitly, and Geoffrey~E Hinton.
\newblock A simple way to initialize recurrent networks of rectified linear
  units.
\newblock {\em arXiv preprint arXiv:1504.00941}, 2015.

\bibitem{kingma2014adam}
Diederik~P Kingma and Jimmy Ba.
\newblock Adam: A method for stochastic optimization.
\newblock {\em arXiv preprint arXiv:1412.6980}, 2014.

\end{thebibliography}
